%% file: ISVLSI_18.tex
\documentclass[conference]{IEEEtran}
\IEEEoverridecommandlockouts
\usepackage{cite}
\usepackage{amsmath,amssymb,amsfonts}
\usepackage{algorithmic}
\usepackage{graphicx}
\usepackage{textcomp}
\usepackage{subcaption}
\usepackage{adjustbox}
\usepackage{balance}
\usepackage{enumitem}
\usepackage{array}
\usepackage{comment}
\def\BibTeX{{\rm B\kern-.05em{\sc i\kern-.025em b}\kern-.08em
    T\kern-.1667em\lower.7ex\hbox{E}\kern-.125emX}}
    
\begin{document}

\title{MAT: A Multi-strength Adversarial Training Method to Mitigate Adversarial Attacks\\
\thanks{This work was supported in part by NSF 1744077 and AFRL ICA2017-UP-018.}
}

\author{\IEEEauthorblockN{Chang Song\IEEEauthorrefmark{1}, 
Hsin-Pai Cheng\IEEEauthorrefmark{1},
Huanrui Yang\IEEEauthorrefmark{1},
Sicheng Li\IEEEauthorrefmark{2}, 
Chunpeng Wu\IEEEauthorrefmark{1}, 
Qing Wu\IEEEauthorrefmark{3},
Yiran Chen\IEEEauthorrefmark{1},
Hai Li\IEEEauthorrefmark{1}}
\IEEEauthorblockA{\IEEEauthorrefmark{1}Department of Electrical and Computer Engineering, Duke University, Durham, NC, USA}
\IEEEauthorblockA{\IEEEauthorrefmark{2}Hewlett Packard Labs, Palo Alto, CA, USA}
\IEEEauthorblockA{\IEEEauthorrefmark{3}Air Force Research Lab, Rome, NY, USA}
\IEEEauthorblockA{\{chang.song, dave.cheng, huanrui.yang\}@duke.edu, sicheng.li@hpe.com, \\chunpeng.wu@duke.edu, qing.wu.2@us.af.mil, \{yiran.chen, hai.li\}@duke.edu}}

\maketitle

\begin{abstract}
Some recent work revealed that deep neural networks (DNNs) are vulnerable to so-called \textit{adversarial attacks} where input examples are intentionally perturbed to fool DNNs.
In this work, we revisit the DNN training process that includes adversarial examples 
into the training dataset so as to improve DNN's resilience to adversarial attacks, namely, \textit{adversarial training}.
Our experiments show that different adversarial strengths, i.e., perturbation levels of adversarial examples, have different working ranges to resist the attacks.
Based on the observation, we propose {a multi-strength adversarial training method} (MAT) that combines the adversarial training examples with different adversarial strengths to defend adversarial attacks.
Two training structures---mixed MAT and parallel MAT---are developed to facilitate the tradeoffs between training time and hardware cost. Our results show that MAT can substantially minimize the accuracy degradation of deep learning systems to adversarial attacks on MNIST, CIFAR-10, CIFAR-100, and SVHN. The tradeoffs between training time, robustness, and hardware cost are also well discussed on a FPGA platform.
\end{abstract}


\vspace{-0.5em}
\input{introduction}
\vspace{-0.5em}
\input{related-work}
\vspace{-0.5em}
\input{motivation}

\vspace{-0.5em}
\input{methodology}
\vspace{-0.5em}
\input{experimental-results}
\vspace{-0.1cm}
\input{conclusions}



\balance
\vspace{-0.1cm}
\bibliographystyle{unsrt}
\bibliography{ISVLSI_18.bib}

\end{document}

%% file: introduction.tex
\section{Introduction}
\label{introduction}

The most exciting advancement of artificial intelligence in the past decade is the wide application of deep learning techniques~\cite{lecun2015deep}.
However, a recent research discovered that machine learning models (including deep neural networks, a.k.a. DNN) are susceptible to \textit{adversarial attacks}, which apply small perturbation on input data to fool the models.
Such attacks normally lead to a lower confidence level or even a misclassification~\cite{szegedy2013intriguing, goodfellow2014explaining}.
In addition, Papernot \textit{et al.} discovered that a perturbed example (i.e., adversarial example) has transferability property: the adversarial example crafted by a model $M_S$ (i.e., substitute model) not only can deceive $M_S$ itself but also can influence other models (i.e., victim models), even without knowing the internal structures of these victim models~\cite{papernot2017arxiva}.
The amplitude of the perturbation that is used in adversarial attacks (a.k.a. \textit{adversarial strength}) can be quite small or even imperceptible to the human eyes. All these properties raise severe concerns on the security of deep learning technique.

In this work, we revisit the recently emerged \textit{adversarial training} process that uses adversarial examples to train DNN and therefore enhances its resilience to adversarial attacks.
We find that the adversarial examples with different adversarial strengths work effectively only for the adversarial attacks with certain range of adversarial strengths.
Based on this observation, we propose a multi-strength adversarial training (MAT) method that defends adversarial attacks by combining adversarial training examples with multiple adversarial strengths.
Two adversarial training structures, namely, mixed MAT and parallel MAT, are developed to integrate the influences of multiple adversarial strengths with different training times and hardware costs. 
The proposed adversarial training structures are also implemented on Xilinx FPGA ZC706 board to evaluate their performances, hardware costs, and energy consumptions and to explore the possible design space.
Compared to the existing works about adversarial attacks and their defense schemes, our major contributions can be summarized as follows:
\begin{itemize}[leftmargin=*]
	\item We identify the limitation on the working range of the existing adversarial training technique against adversarial attacks;
	\item We invent multi-strength adversarial training  (MAT)---the first adversarial training method that can enhance the resilience of learning systems over a controllable wide range of adversarial length under adversarial attacks;
	\item We propose mixed MAT and parallel MAT to facilitate a flexible tradeoff between training time and hardware cost;
	\item We implement a random walk algorithm to optimize the selections of adversarial strengths and other design parameters in the MAT process; and
    \item We also implement MATs with different configurations on a FPGA platform and discuss the tradeoffs between training time, robustness, and hardware cost.
\end{itemize}

Our experimental results on MNIST, CIFAR-10, CIFAR-100, and SVHN 
show that both mixed MAT and parallel MAT can better defend the learning model under adversarial attacks than single-strength adversarial training and parallel MAT offers the largest accuracy improvement. 
The results also show that the model robustness is greatly affected by the complexity and size of the network structure, the training dataset, the associated hardware cost, and training time. 

The remainder of this paper is organized as follows: Section~\ref{related-work} gives preliminary about adversarial attacks and its defense techniques; Section~\ref{motivation} introduces the motivation of our work; Section~\ref{methodology} presents the details of our proposed method; Section~\ref{experimental-results} shows the experimental results and discussions; Section~\ref{conclusions} concludes this work. 

%% file: related-work.tex
\section{Preliminary}
\label{related-work}

A robust learning model is expected to be able to tolerate random noises in input samples to certain degree~\cite{Ulicny2016robustness}.
However, recent studies showed that the robustness of a learning model is threatened by so called \textit{adversarial attacks}~\cite{papernot2017practical}.
To be specific, a learning model may misclassify an example that is carefully perturbed, say, an \textit{adversarial example}, to a wrong class.
An adversarial example $\tilde{X}$ can be generated by injecting perturbation $\varepsilon$ to the original input sample $X$ such as $\tilde{X}=X+\varepsilon$.
The linear transformation of $\tilde{X}$ with respect to a given weight vector $W$ is
\begin{equation}
\label{eq:intro}
W^{T}\tilde{X}=W^{T}X+W^{T}\varepsilon.
\end{equation}
Here the adversarial perturbation $\varepsilon$ is often referred to as adversarial strength.
$W^{T}\tilde{X}$ increases proportionally with the dimensionality of $W$.
Since $W$ is often high dimensional in practical problems, a minor perturbation $\varepsilon$ could introduce a big change of $W^{T}\tilde{X}$.
Such a small change may not be caught visually or by a rule-based detection scheme but could be sufficient to lead a misclassification in DNNs.

Some popular methods to defend adversarial attacks are:

\begin{itemize}[leftmargin=*]
	\item \textit{Gradient masking:} Its main idea is to build a model to hide or smooth the gradient between original and adversarial examples~\cite{papernot2017practical}.
	The effectiveness of this method, however, can significantly degrade when the attacker uses a model different from the protected model to generate the adversarial examples.
	\item \textit{Defensive distillation:} The target of this method is to create a model whose decision boundaries are smoothed along the directions that the attacker may exploit. Defensive distillation makes it difficult for the attacker to discover adversarial input tweaks that lead to incorrect classes.
	\item \textit{Adversarial training:} The objective of adversarial training is using adversarial examples to train the model and therefore enhance its resilience to adversarial attacks. Its effectiveness has been shown and explained by Goodfellow \textit{et al.}~\cite{goodfellow2014explaining}. Note that in adversarial training, the model that is used to generate the adversarial examples is not necessarily identical to the model being attacked. 
\end{itemize}

%
In some recent relevant research works, Kurakin \emph{et al.}~\cite{Kurakin2017ICLR} show that combining small batches of both adversarial examples and original data in adversarial training could make the model more resilient to adversarial attacks. 
Carlini and Wagner ~\cite{carlini2017towards} demonstrate that defensive distillation does not significantly enhance the robustness of neural networks in some scenarios by introducing three new attack algorithms. 
Cisse \emph{et al.}~\cite{cisse2017parseval} introduce a layer-wise
regularization method to reduce the neural network’s sensitivity to small perturbations, which are difficult to be visually caught.
However, these work do not give experimental guidance on how to adaptively select the adversarial strength of the adversarial examples that are used in adversarial trainings or attacks to maximize either the defense or attack effectiveness. And to the authors' best knowledge, there is no work discusses the implementation on hardware and hardware consumption.

%% file: motivation.tex
\section{Motivation}
\label{motivation}

\begin{figure}[t]
	\centering
	\includegraphics[width=\linewidth]{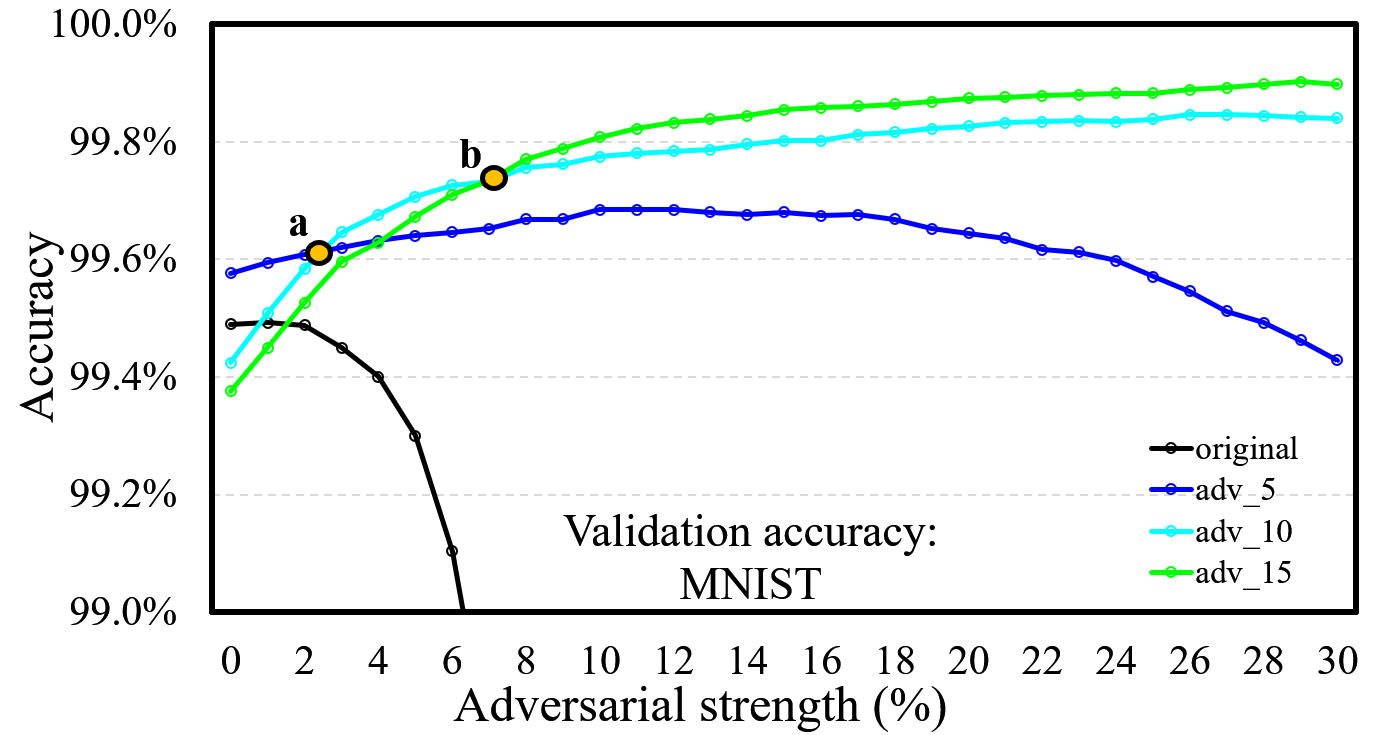}
	\caption{Robustness of the models trained with different adversarial strengths.}
	\label{fig:Accuracy_drop}
\end{figure}

It is known that training the DNN using adversarial examples with certain adversarial strength helps in improving the DNN’s resilience against adversarial attacks.
Figure~\ref{fig:Accuracy_drop} compares the robustness (overall accuracy performance in the interested adversarial strength range) of a 6-layer neural network trained with different datasets on MNIST.
Here the \textit{original} model is trained with the original dataset, and the models of \textit{adv\_5}, \textit{adv\_10}, and \textit{adv\_15} are trained with half legitimate examples and half adversarial examples with the adversarial strength = 0.05, 0.10, and 0.15, respectively.
As we can see from the results, the model trained with the original dataset is very susceptible to the increase of adversarial strength during the adversarial attacks.
However, including the adversarial examples in the training process can effectively maintain the model's accuracy when the adversarial strength increases, which aligns with the result in Goodfellow \textit{et al.}~\cite{goodfellow2014explaining}.

In addition, Figure~\ref{fig:Accuracy_drop} shows that the accuracies of the models trained with different adversarial strengths cross each other over the simulated range of the adversarial strengths adopted in the attacks.
As we can see from the figure, these models demonstrate different defending effectiveness on different adversarial strength ranges.
On the left side of point \textbf{a}, for example,
the model \textit{adv\_5} demonstrates the highest accuracy among all the trained models. 
As the adversarial strength adopted in the attacks increases, the models \textit{adv\_10} and \textit{adv\_15} give the highest accuracy in turn. 
This observation in fact reveals a limitation of the single strength adversarial training: each adversarial strength of the samples used in the adversarial training has its best working range, say, around the same strength that is adopted in the attacks.

By leveraging the different working ranges of the different adversarial strengths used in adversarial training, there exists a possibility to develop a new adversarial training method that can combine these working ranges so that the trained learning system can be resilient to the attacks over a wide range of the adversarial strength.

\begin{figure}[t]
	\centering
	\includegraphics[width=\linewidth]{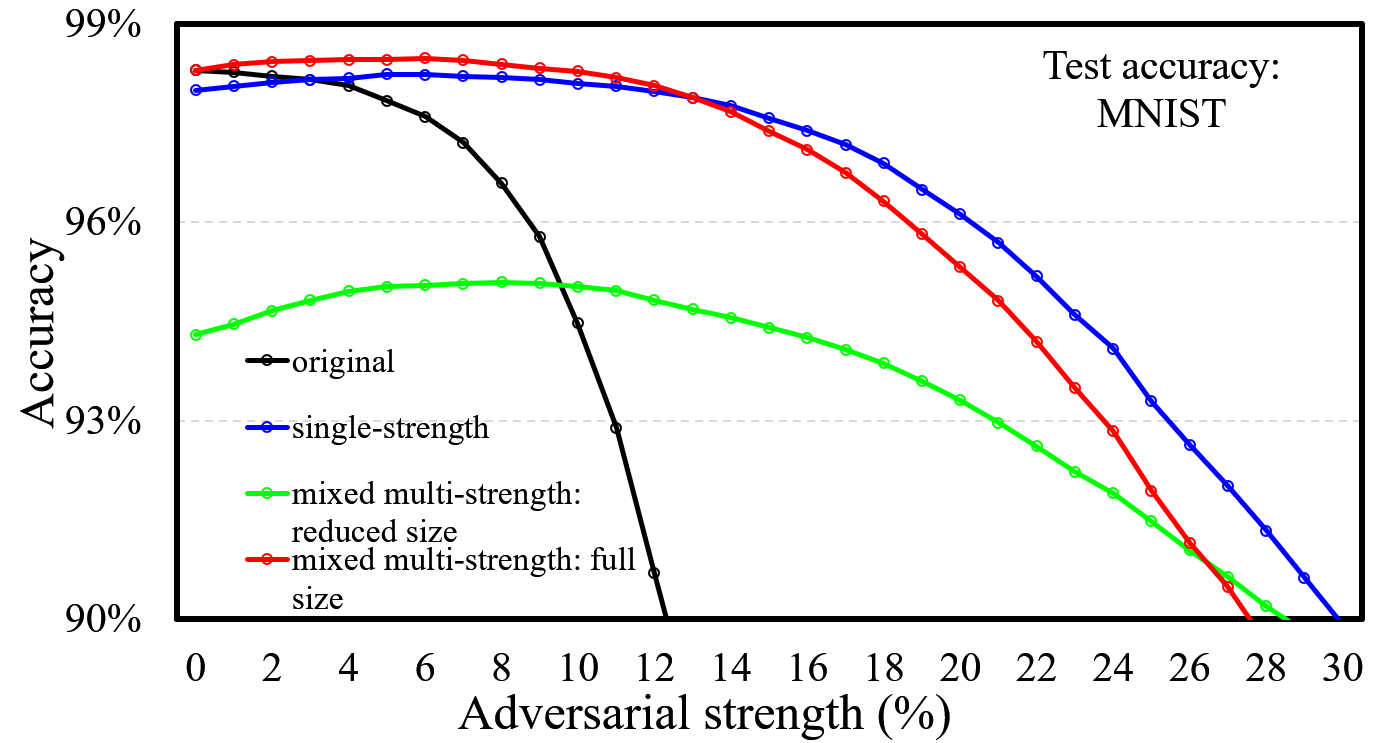}
	\caption{Robustness comparison between the original, the single-strength adversarial, and the mixed-strength adversarial training.}
	\label{fig:comparison_training_methods}
\end{figure}

Figure~\ref{fig:comparison_training_methods} shows our initial simulation results of the accuracy of the models adversarially trained with different configurations of the training dataset under the adversarial attacks with different strengths.
Similar to Figure~\ref{fig:Accuracy_drop}, here model \textit{original} represents our baseline, which is trained with the original dataset. Model \textit{single-strength} is trained with half of the original dataset and half of the adversarial examples generated by Fast Gradient Sign Method (FGSM)~\cite{goodfellow2014explaining}.
Model \textit{mixed multi-strength: reduced size} is trained with a mixed combination of the original dataset and the adversarial dataset with the strengths of 0.05, 0.10, and 0.15, respectively. 
The size of each subset of the training data is 25\% of the original dataset size. 
Model \textit{mixed multi-strength: full size} has the same partition of all datasets as \textit{mixed multi-strength: reduced size} does, but the size of each subset of the data is the same as that of the original dataset.
Our simulation results show that model \textit{mixed multi-strength: full size} performs the best over a considerably large range of the adversarial strength, indicating a great potential of our proposal to combine the adversarial examples with different strengths for robust model training.
As we shall show later, the selections of the adversarial strength levels and the number of strengths are critical in our proposal.
The results also show that packing the data with different adversarial strengths into the same size of the original dataset, i.e., as model \textit{mixed multi-strength: reduced size}, may not help much to enhance the model's resistance and may even degrade the model accuracy. 
A possible mathematic explanation about why combining multiple adversarial strengths during the adversarial training helps in improving the resilience of the model to the adversarial attacks is related to the construction of the decision hyperspace during the training. Due to space limit, we do not include this explanation in this paper.

%% file: methodology.tex
\section{Methodology}
\label{methodology}

In light of the limited working range of single-strength adversarial training, in this work, we propose multi-strength adversarial training (MAT) 
to combine the effects of multiple adversarial strengths to improve the robustness of the neural network over a wide adversarial strength range under adversarial attack.
Adversarial examples with different adversarial strengths are mixed with the original training dataset in MAT.
The total size of the new training dataset $N$ for MAT, hence, becomes \(N=N_C+S \cdot N_A\).
Here $N_C$ is the size of the original training dataset. $N_A$ is the size of the generated adversarial dataset with a certain adversarial strength.
$S$ denotes the number of the different adversarial strengths that are adopted in MAT.

Two training structures, namely, mixed multi-strength adversarial training (mixed MAT) and parallel multi-strength adversarial training (parallel MAT), are proposed to facilitate the tradeoff between the training time and the hardware cost of MAT.
Some automated optimization method, e.g., one-dimensional random walk algorithm, can be used to select the optimum adversarial strengths adopted in MAT.
The details of these techniques will be explained in this section.

\vspace{-0.5em}
\subsection{Mixed MAT}

\begin{figure}[b]
	\centering
	\begin{subfigure}{0.25\linewidth}
		\centering
		\includegraphics[height=5.15cm]{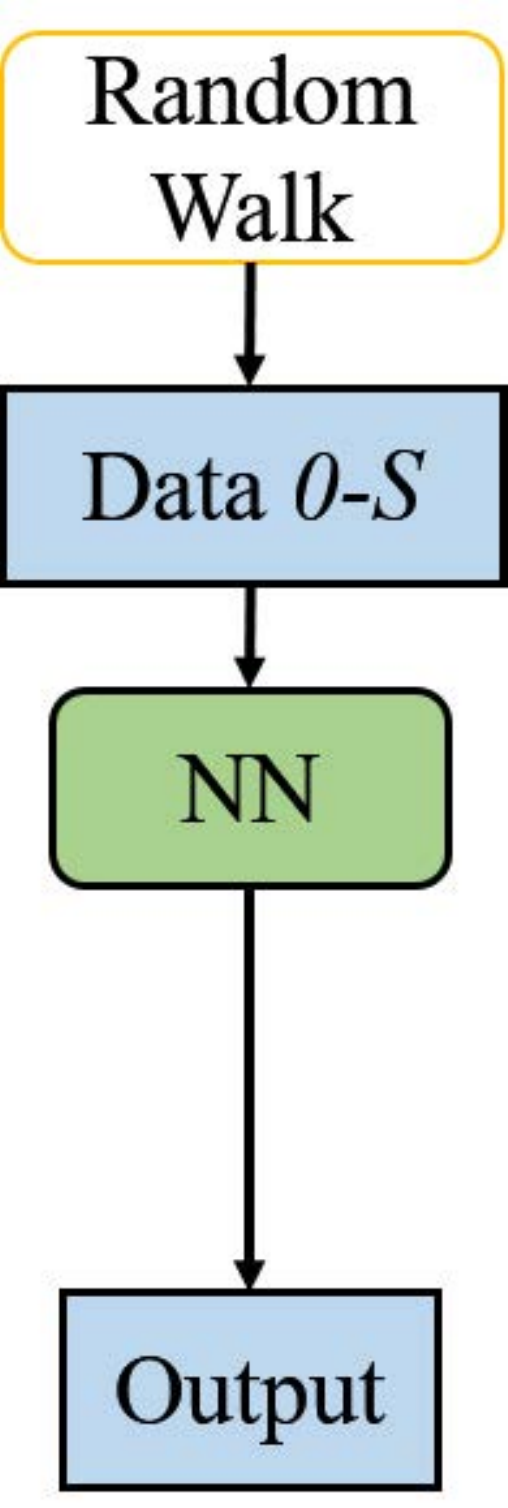}
		\caption{Mixed MAT}
		\label{fig:MMAT}
	\end{subfigure}
	\hfil
	\begin{subfigure}{0.7\linewidth}
		\centering
		\includegraphics[height=5cm]{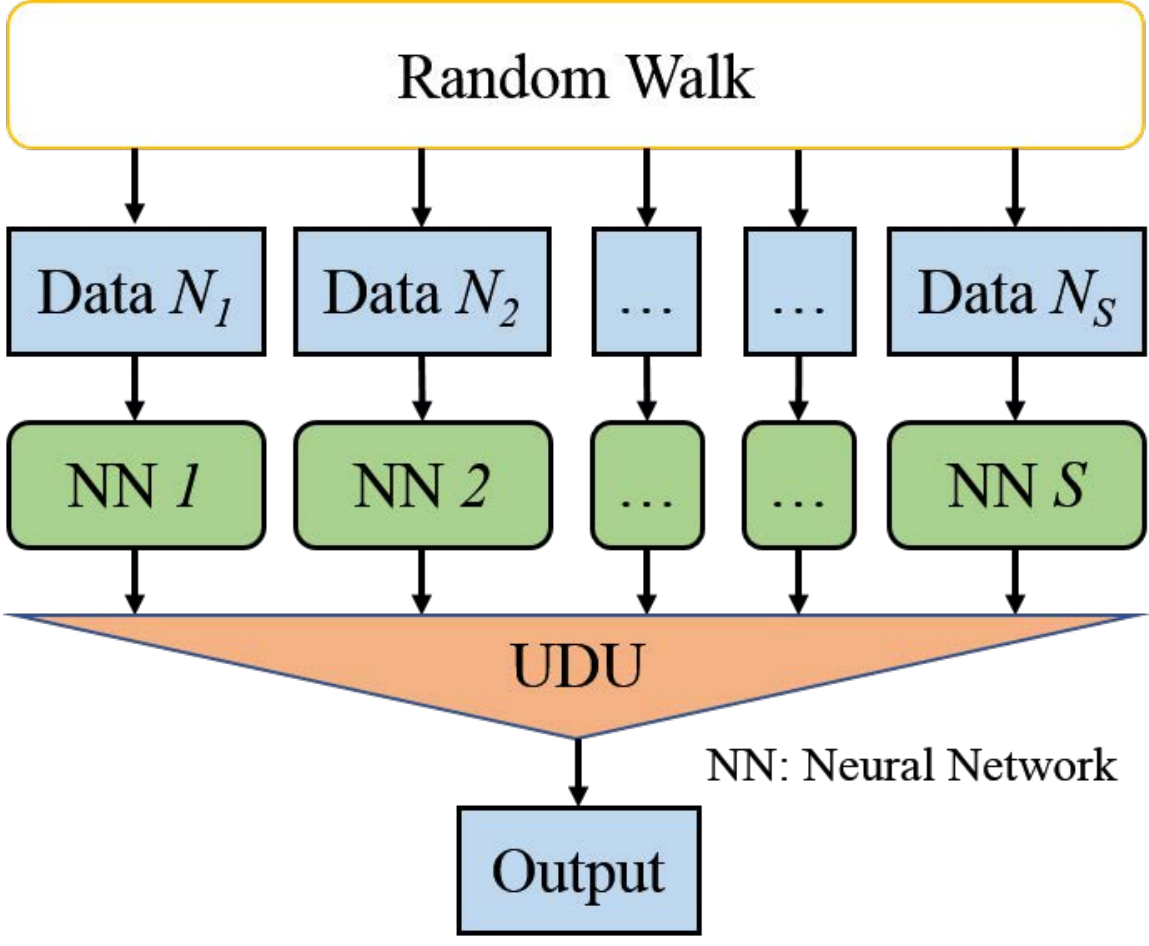}
		\caption{Parallel MAT}
		\label{fig:PMAT}
	\end{subfigure}
	\caption{Our proposed structures of mixed MAT and parallel MAT.}
	\label{fig:MATs}
\end{figure}

Figure~\ref{fig:MATs}(\subref{fig:MMAT}) illustrates the training structure of mixed MAT.
The new training dataset $N$, which includes the original training dataset and $S$ generated adversarial datasets, are numbered from $0$ to $S$ as the training input of the neural network.
Here number $0$ represents the original training dataset.
A modified loss function of mixed MAT can be constructed as:
\begin{equation}
\label{eq:1}
J_M=\sum_{i=1}^{N_C}J(\boldsymbol\theta, X_i, T_i)+\sum_{j=1}^{S}\sum_{i=1}^{N_A}J(\boldsymbol\theta, f(X_i), T_i).
\end{equation}
Here $X_i$ is the $i$-th original example;
$T_i$ stands for the $i$-th target value;
$f(\cdot)$ is the function of adversarial transformation.
The new loss function $J_M$ contains two terms, which respectively represent the original training part and the newly-added adversarial training part with total $S$ different adversarial strengths.
The interior sum of the second term denotes the loss on every single strength adversarial dataset, while the exterior sum denotes the overall loss of all the adversarial datasets with different adversarial strengths.

Assume that the sizes of the original and the adversarial dataset are identical, then the total training time of mixed MAT will be $(S+1)$ times of the original training time of the neural network.
During practical applications of mixed MAT, however, the size of the used datasets may be reduced from that of the original dataset to save the training time, by paying the cost of possible model accuracy degradation.
Nonetheless, mixed MAT affects neither the computational complexity nor the execution time of the testing process of the neural network.
The network structure is not changed either.

\vspace{-0.5em}
\subsection{Parallel MAT}

We need to point out that directly combining the adversarial datasets with different adversarial strengths, i.e., in mixed MAT, is not the only option to leverage the different working ranges of these datasets.
A more straightforward thinking is that for a specific range of the adversarial strengths adopted by an adversarial attack, we shall always train the neural network with the adversarial dataset that ensures the highest model accuracy under the attack, as illustrated in Figure~\ref{fig:Accuracy_drop}.
Following this philosophy, we propose parallel MAT, the concept of which is illustrated in Figure~\ref{fig:MATs}(\subref{fig:PMAT}).

In parallel MAT, total $S$ neural network copies can be trained in parallel.
Each of these copies is trained with the combination of the original dataset and one of the adversarial datasets with a certain adversarial strength.
The overall size of the new training dataset, hence, becomes $2S$ times of the original dataset.
Accordingly, the loss function of the $k$-th ($k=1, 2, \ldots, S$) neural network copy is modified to:
\begin{equation}
\label{eq:2}
J_{P,k}=\sum_{i=1}^{N_C}J(\boldsymbol\theta, X_i, T_i)+\sum_{i=1}^{N_A}J(\boldsymbol\theta, f(X_i), T_i).
\end{equation}
Different from Eq.~(\ref{eq:1}) where the effects of the original dataset and all the adversarial datasets are taken into account as a whole, Eq.~(\ref{eq:2}) particularly focuses on the robustness enhancement of the neural network over the working range of a single adversarial dataset.
The outputs (e.g., the loss functions or the predicted possibilities) of all the neural copies are then summarized in an upper-boundary decision unit (UDU), as shown in Figure~\ref{fig:MATs}(\subref{fig:PMAT}).
The function of UDU $v(\cdot)$ is to collect the outputs from all the DNN copies and then decide the classification result by a voting process such as:
\begin{equation}
\label{eq:3}
J_P=\sum_{k=1}^{S}a_k v(J_{P,k}),
\end{equation}
where $a_k$ is the coefficient of voting for the $k$-th neural network copy $v(J_{P,k})$ and satisfies $\sum_{k=1}^{S} a_k=1$.
In the implementation of parallel MAT, $a_k$ can be learned using a shallow neural network.

Compared to mixed MAT, parallel MAT reduces the total training time of the DNN by leveraging the computation parallelism.
However, the hardware implementation cost may significantly increase in parallel MAT by replicating the neural network. 
We note that in practice, the optimal numbers of adversarial strengths adopted by mixed MAT and parallel MAT are not necessarily the same.

\vspace{-0.5em}
\subsection{Multi-strength Selection}

We utilize one-dimensional random walk algorithm~\cite{harel2007graph} to automatically select the adversarial strengths adopted in MATs.
A random walk is a stochastic process that describes a path formed by a succession of random steps on some mathematical space.
The one-dimensional random walk algorithm used in this paper includes the following three procedures:

\begin{itemize}[leftmargin=*]
\item \textit{Pre-computation}:
An accuracy matrix $A$ that contains the average single-strength training accuracies in an adversarial strength range is measured in the victim model and provided to the random walk function.
Here we use \textit{validation accuracy} to approximate test accuracy by assuming that the victim model cannot access the test dataset.
	
\item \textit{Initialization}:
Based on the accuracy matrix $A$, the multi-strength accuracy matrix $A_M$ can be estimated as $A_M(i, j)=A(i, j)/\sum(A(i))$, where $\sum(A(i))$ denotes the sum of the $i$-th row of matrix $A$.
According to $A_M$, we can calculate the state transition matrix $P(A)$ when walking from one single-strength state to another and initialize the multi-strength accuracy estimation function $H(\cdot)$.

\item \textit{Simulation}:
During the iterative simulation, we perform $l$-step random walk for $t$ times to estimate $H(A_M, l, t)$.
To limit the total number $S$ of the selected adversarial strengths, a penalty term with an coefficient $a$ is added to calculate the multi-strength accuracy estimation by using random walk function, such as $G(A, l, t) = H(A_M, l, t) + a \cdot S$.
\end{itemize}

After exercising sufficient steps of random walk, $G(\cdot)$ will give the best estimated accuracy that corresponds to an optimal combination of multiple adversarial strengths represented by $A_M$.
Such a selection method can be used in both mixed MAT and parallel MAT, though they might start with different accuracy matrices.

%% file: experimental-results.tex
\section{Experimental results}
\label{experimental-results}

In this section, we compare mixed MAT and parallel MAT with the single-strength adversarial training on four image datasets: MNIST, CIFAR-10, CIFAR-100, and SVHN. 
MNIST consists of digits, and CIFAR-10 and CIFAR-100 consists of natural scenes with different class numbers.
SVHN is similar to MNIST where each image is a street view house number.
In addition, we also implement these adversarial training schemes with different configurations and discuss the relevant tradeoffs.

\vspace{-0.5em}
\subsection{Experimental Setup}

We use Fast Gradient Sign Method (FGSM) to craft both training and testing datasets.
The detailed model structure and setting are described as follows:

\textbf{Upper bound of adversarial strength:}
We use Mean Structural SIMilarity (MSSIM) index~\cite{wang2004image} to limit the maximum value of the adversarial strength $\varepsilon$.
MSSIM is a value between -1 and 1 to measure the similarity between two images and can be also used to describe the distortion of an image.
In this work, we set lower bound of the MSSIM between 0.77 and 0.82, which corresponds to a distortion that can be visually captured.

\textbf{MNIST:}
We use LeNet-5 with an accuracy of 99.5\% to craft adversarial examples.
The range of adversarial strengths $\varepsilon$ is determined upon the following observation:
by testing adversarial examples with different adversarial strengths, we find that as $\varepsilon$ increases from 0 to 0.09, 0.15 and 0.30, the corresponding MSSIM decreases from 1.00 to 0.94, 0.90 and 0.82.
It implies $\varepsilon \in [0, 0.3]$ will be enough for our evaluation. Similar criteria are applied to other datasets.

\textbf{CIFAR-10, CIFAR-100, and SVHN:}
CIFAR-10 and CIFAR-100 use the same model to craft adversarial examples.
The model contains 3 convolutional layers and 2 fully connected layers.
Each convolutional layer is followed by a batch normalization layer and a max pooling layer.
The original accuracies of CIFAR-10 and CIFAR-100 are 82.7\% and 54.3\%, respectively.
The adversarial CIFAR-10 and CIFAR-100 examples are generated in the adversarial strength range of $\varepsilon \in [0, 5]$, with a step size of 0.5.
SVHN follows the same procedure as that of CIFAR-10 and CIFAR-100 in crafting adversarial examples.


For victim model, we use a 3-layer multilayer perceptron (MLP) for MNIST dataset, a convolutional neural network (CNN) with 3 convolutional layers 
for CIFAR-10, CIFAR-100, 
and SVHN. 
Here no data augmentation method (e.g., cropping or mirroring) is used in the training process. We test every method on full adversarial test dataset 
in the concerned adversarial strength range as aforementioned; In every adversarial attack, we assume the test examples are perturbation with the same adversarial strength. 

\vspace{-0.5em}
\subsection{Technology Comparisons}

\begin{figure}[t]
	\centering
	\begin{subfigure}{0.4\textwidth}
		\centering
		\includegraphics[width=\textwidth]{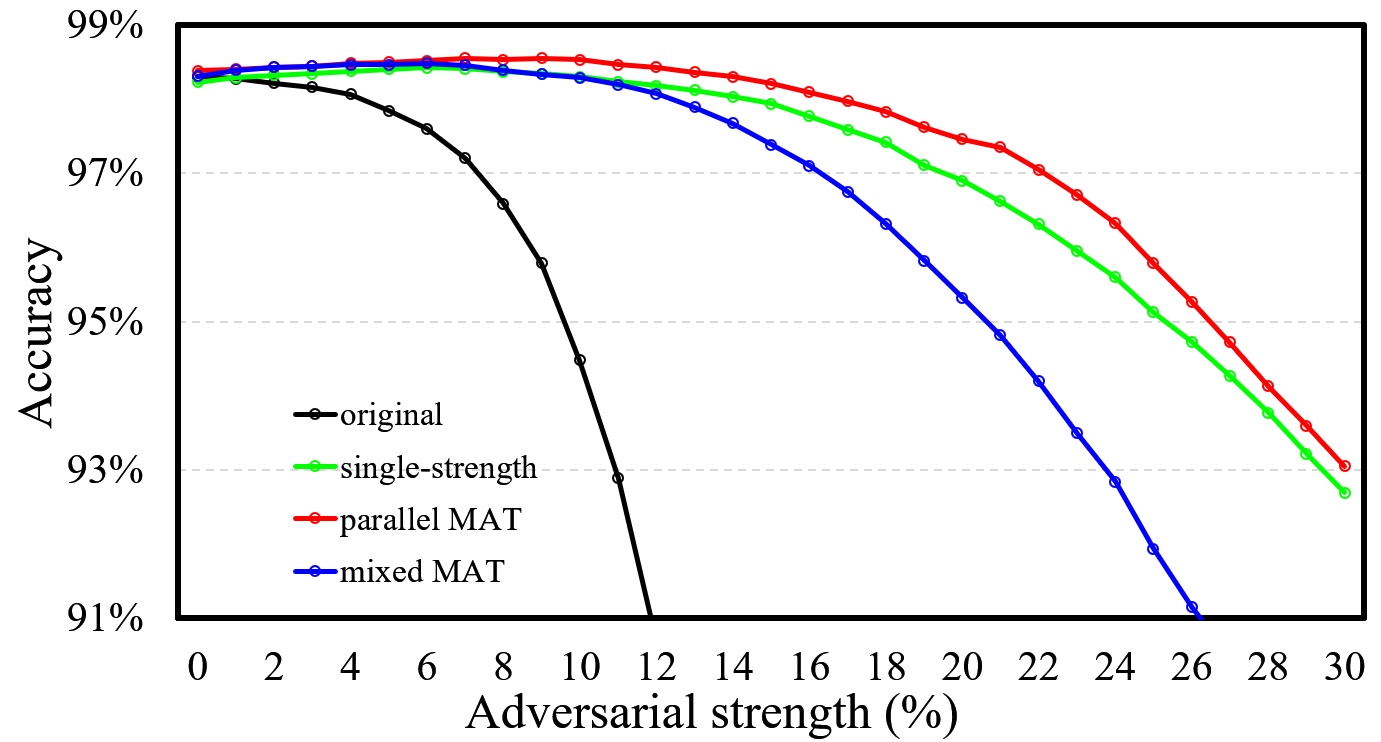}
		\caption{MNIST}
		\label{fig:MNIST}
	\end{subfigure}
	\begin{subfigure}{0.4\textwidth}
		\centering
		\includegraphics[width=\textwidth]{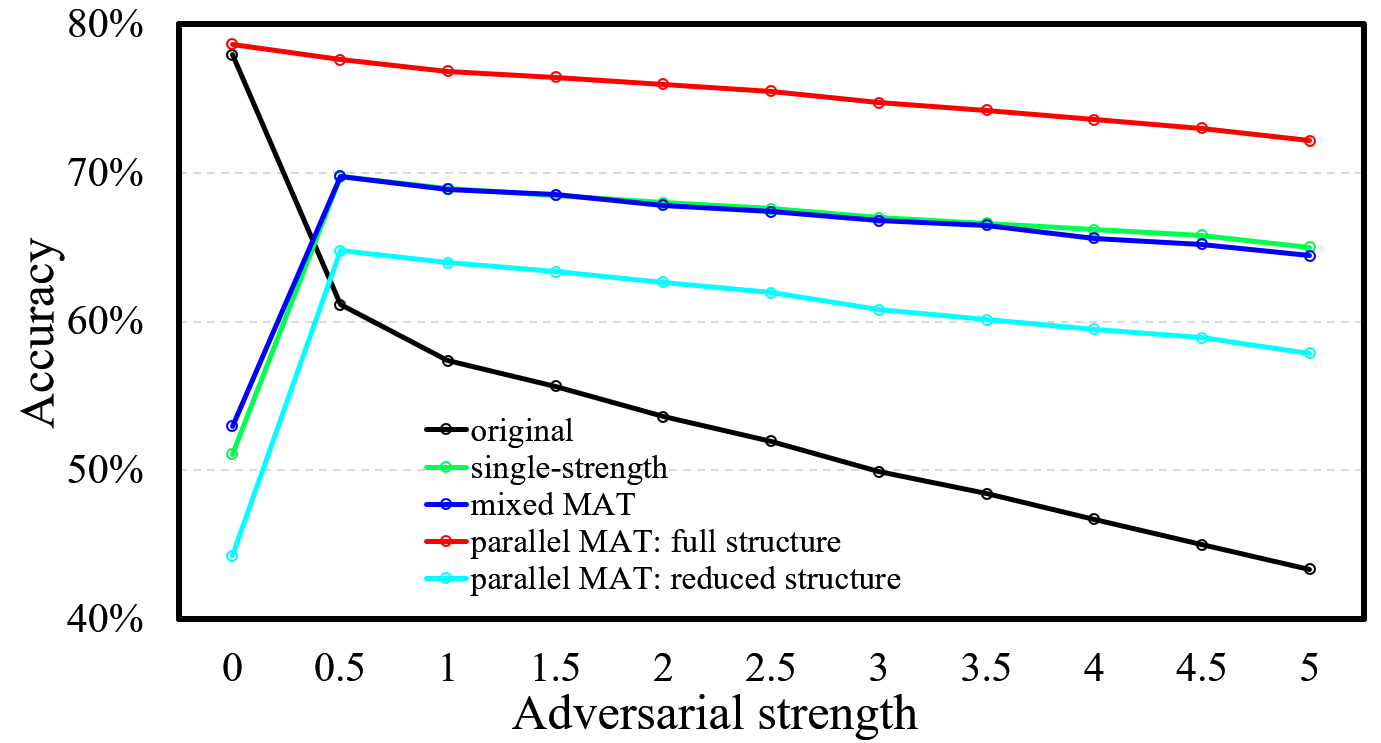}
		\caption{CIFAR-10}
		\label{fig:CIFAR-10}
	\end{subfigure}
	\begin{subfigure}{0.4\textwidth}
		\centering
		\includegraphics[width=\textwidth]{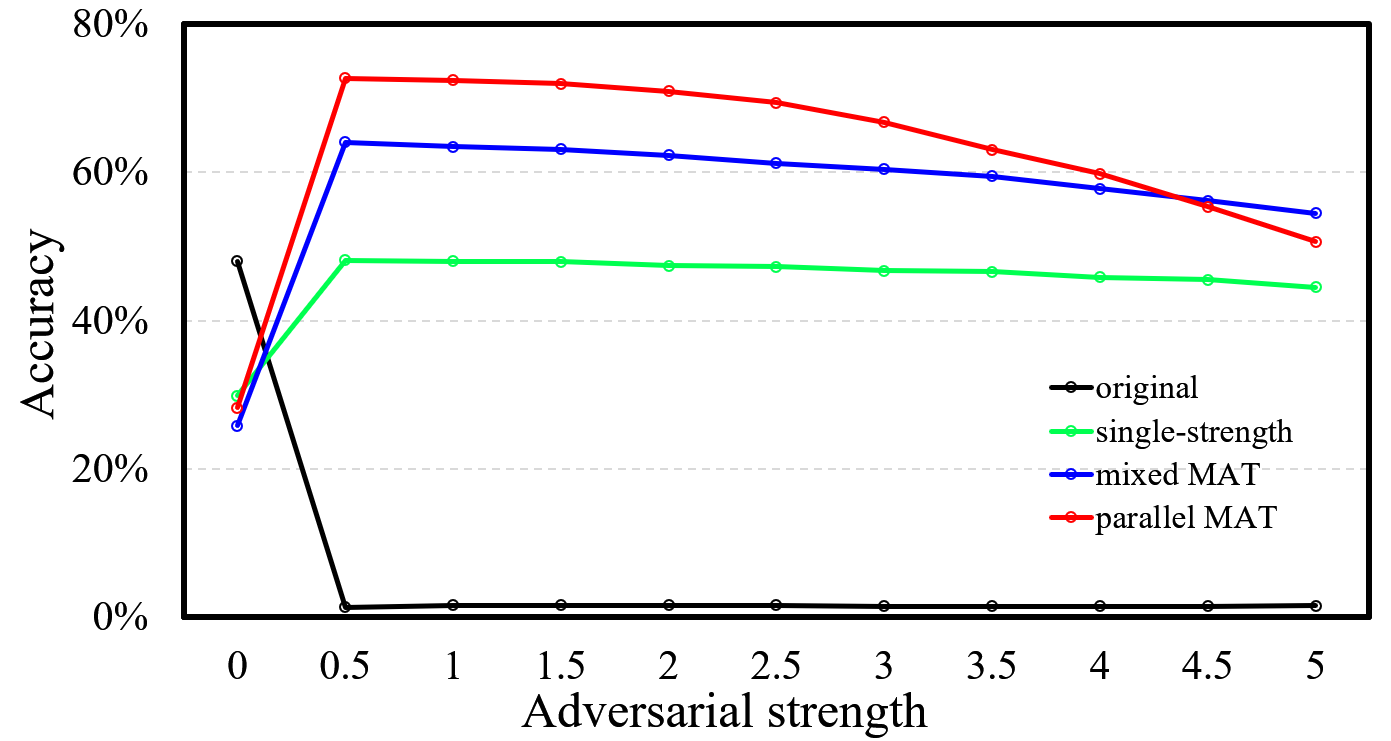}
		\caption{CIFAR-100}
		\label{fig:CIFAR-100}
	\end{subfigure}
	\begin{subfigure}{0.4\textwidth}
		\centering
		\includegraphics[width=\textwidth]{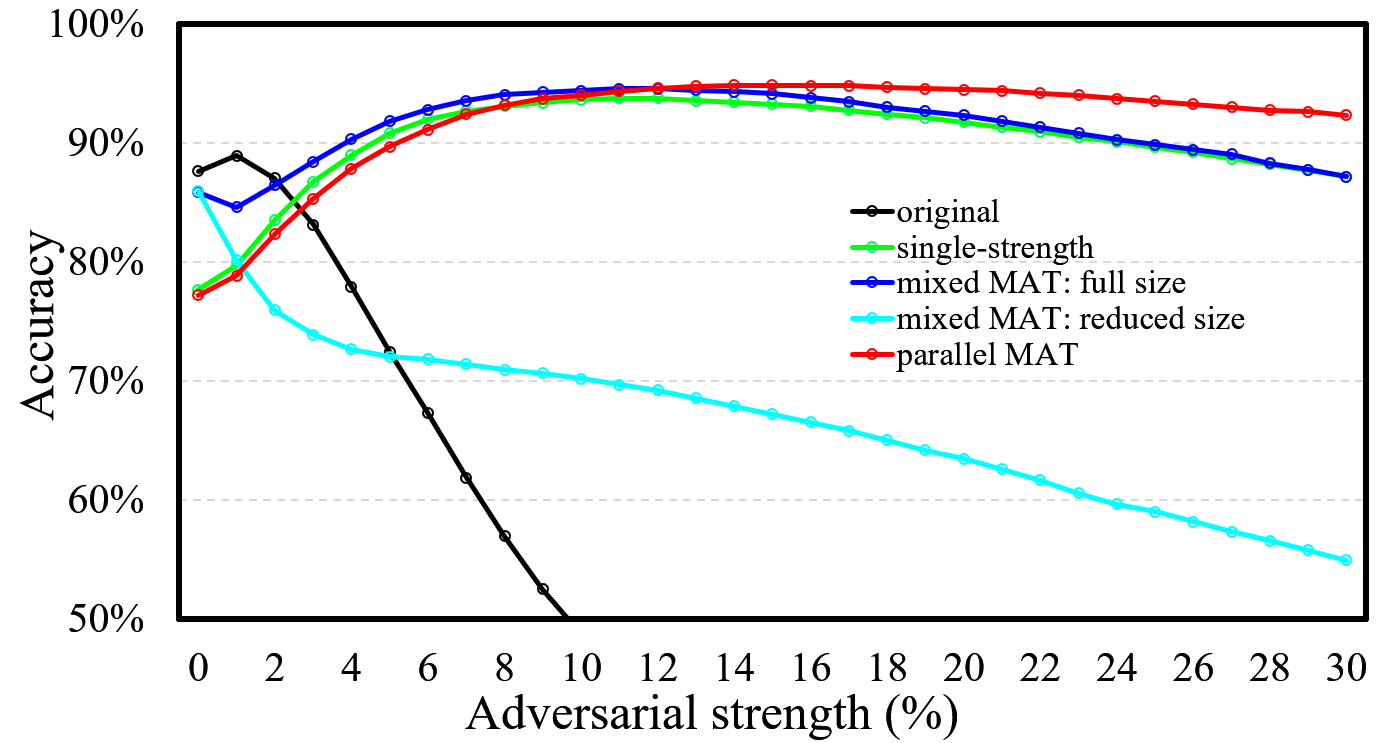}
		\caption{SVHN}
		\label{fig:SVHN}
	\end{subfigure}
	\caption{Robustness of proposed MATs on different datasets.}
	\label{fig:results}
\end{figure}

Figure~\ref{fig:results} compares the effectiveness of four training schemes on enhancing the model's resilience to adversarial attacks, including the original data training, the single-strength adversarial training, and our proposed mixed MAT and parallel MAT.
For each sub-figure, the horizontal axis is the testing adversarial strength, and the vertical axis is the model's testing accuracy.
Each of them corresponds to one dataset, i.e., (a) MNIST, (b) CIFAR-10, (c) CIFAR-100, and (d) SVHN. 


\textbf{Single-strength vs. Original.}
Single-strength adversarial training achieves a better accuracy performance than the original data training on adversarial examples, but has a lower accuracy on the original examples.
Adversarial perturbation can be understood as a specified distortion of the original examples.
Introducing random distortions or other data augmentation methods in the training examples usually improves the training accuracy because of the expansion of sample spaces.
However, in single-strength adversarial training, the adversarial examples push the decision boundary towards other classes and may lead to misclassification of some of the originally correctly-classified samples. 

\textbf{Mixed/Parallel MAT vs. Single-strength.}
For all datasets, parallel MAT generally achieves the highest accuracy over the simulated adversarial strength range, and outperforms single-strength adversarial training.
Mixed MAT, however, demonstrates a higher accuracy than single-strength adversarial training only over limited range of the adversarial strength on most datasets.
Both parallel and mixed MAT select the adversarial strengths using random walk. 
The reason that parallel MAT substantially outperforms mixed MAT is possibly because the network structure of mixed MAT is not capable of learning original and adversarial information very well simultaneously. 

\vspace{-0.5em}
\subsection{Performances on Different Datasets}

MNIST and CIFAR-10 have similar amount of training and testing examples but CIFAR-10 is more complex than MNIST as CIFAR-10 has RGB channels.
As shown in Figure~\ref{fig:results}(\subref{fig:MNIST}), for MNIST, single-strength adversarial training is able to compensate the accuracy loss decently.
For CIFAR-10, the baseline accuracy of the model trained with the original data is 77.96\%.
Following the increase of adversarial strength, the model accuracy drops steeply, e.g., down to 43.34\% when $\varepsilon$ is merely 5.
However, when single-strength adversarial training is applied, the accuracy is restored back to 65\%$\sim$70\% within the simulated $\varepsilon$ range.
For SVHN, both MATs demonstrate very impressive capability to enhance the model's resilience to adversarial attacks. The relevant results are summarized in Table~\ref{tab:results}.


\begin{table}[t]
	\centering
	\caption{Average test accuracy in the interested strength range on different datasets.}
	\begin{tabular*}{\linewidth}{m{2.4cm}||m{1cm}|m{1cm}|m{1cm}|m{1cm}}
    	\hline
		Training Methods & Original & Single-strength & Mixed MAT & Parallel MAT \\
		\hline \hline
		\textit{CIFAR-10} & 53.73\% & 65.86\% & 65.81\% & \textbf{75.35\%}  \\
		\textit{CIFAR-100} & 5.77\% & 45.27\% & 57.12\% & \textbf{61.94\%}  \\
		\textit{MNIST} & 82.13\% & 96.50\% & 95.76\% & \textbf{97.36\%}  \\
		\textit{SVHN} & 44.04\% & 90.21\% & 91.31\% & \textbf{91.84\%}
       \tabularnewline
		\hline 
	\end{tabular*}
	\label{tab:results}
\end{table}



\vspace{-0.5em}
\subsection{Design Exploration}
We can reduce the cost of the parallel MAT down to the same level as mixed MAT has by using simplified network structure on each network copy.
The accuracy result of one example of such a reduced structure is presented in Figure~\ref{fig:results}(\subref{fig:CIFAR-10}) as ``parallel MAT: reduced structure''.
As can be seen from the figure, the parallel MAT with reduced-structure can still achieve an overall higher accuracy than the original training, but it fails to match the results of the two MATs (with full structures). This result implies a possible tradeoff between accuracy and hardware cost in parallel MAT design, which will be further discussed in Section~\ref{FPGA}. 

Moreover, Figure~\ref{fig:results}(\subref{fig:SVHN}) (SVHN) shows that mixed MAT with reduced training data size 
shows much worse accuracy than mixed MAT with the full training data size, which echoes the result of Figure~\ref{fig:comparison_training_methods} in Section~\ref{motivation}.
We note that for parallel MAT, it is not sufficient for each network copies to learn enough information if we further reduce the training data size. 
Therefore, this option is beyond our consideration in design exploration.

\vspace{-0.5em}
\subsection{Implementation on FPGA Platforms}
\label{FPGA}
\begin{table}
\vspace{-1em}
\caption{Resource utilization of Mixed/Parallel/Structure-reduced Parallel MAT on FPGA, using CIFAR-10 dataset.}
	\label{tab:hardware}
\tabcolsep 2pt
\begin{tabular*}{\linewidth}{@{\extracolsep{\fill}}c||c|c|cccc}
\hline 
 & MACC  & Model  & \multicolumn{4}{c}{Hardware Resource }\tabularnewline
 & \# of Operations  & Size & LUT & FF & BRAM & DSP\tabularnewline
\hline 
\hline 
Mixed MAT & 12.3M & 89.4k & 27761 & 26600 & 75 & 220\tabularnewline
\hline 
Parallel MAT & 49.3M & 360.3k & 139385 & 85172 & 390 & 900\tabularnewline
\hline 
Reduced & 10.9M & 57.1k & 43118 & 34097 & 203 & 400\tabularnewline
\hline 
\end{tabular*}
\end{table}

To give a more specific understanding on different models, we evaluate the hardware implementation cost of different designs based on their corresponding FPGA realizations that are designed with Vivado HLS 2016.4. 
This tool initializes the implementation with C language and then exports the RTL as an IP core. 
Fast C/RTL co-simulation is used for design space exploration and hardware resource estimation. 
The design is deployed on a single FPGA and uses DRAM as external storage. 
We use systolic arrays of uniformed processing elements (PEs) as the main computing units with 32-bit floating-point precision. 
The global control unit initializes the accelerator and distributes kernel weights and feature maps to PEs at runtime. 
The data from/to the external memory is handled by a multi-port DMA streaming engine. 
Each PE is assigned with a subset of the overall computation, the PE controller sets up registers according to the received configuration instructions, and then enables Data Fetcher to load vector arrays of an input feature map into on-chip buffer at runtime. 
The PE also integrates ReLU activation and Pooling function.
After placement and routing, the chip operates at 150 MHz. 

Table~\ref{tab:hardware} summarizes the resource utilization of different network structures trained with CIFAR-10 on Xilinx ZC706 development board.  ``Reduced'' in the table indicates ``parallel MAT: reduced structure'' as in Figure~\ref{fig:results}(\subref{fig:CIFAR-10}). The comparison between different models in terms of accuracy and hardware cost shows that the bigger model size and higher computation density the model has, the more robust it will be. But the training time may not follow this rule because of the computation parallelism. These results could guide us to the tradeoff between robustness, training time, and hardware cost, which can facilitate the hardware designs in the future. 

%% file: conclusions.tex
\section{Conclusions}
\label{conclusions}

In this work, we observe that single-strength adversarial training demonstrates limited working range to enhance the model's resilience to adversarial attacks. 
Hence, we propose two multi-strength adversarial training (MAT) methods, namely, mixed MAT and parallel MAT, to alleviate adversarial attacks.
Moreover, a random walk algorithm is adopted to optimize the selections of the adversarial strengths that are included in the two MAT methods.
Our experimental results on four different datasets 
show that compared to the single-strength adversarial training method, both mixed MAT and parallel MAT substantially improve DNN model's resilience to adversarial attacks. 
The results also indicate that higher robustness can be achieved by higher computation density and bigger model size, but the training time can be greatly reduced by using computation parallelism on a FPGA platform.

%% file: ISVLSI_18.bbl
\begin{thebibliography}{10}

\bibitem{lecun2015deep}
Yann LeCun, Yoshua Bengio, and Geoffrey Hinton.
\newblock Deep learning.
\newblock {\em Nature}, 521(7553):436--444, 2015.

\bibitem{szegedy2013intriguing}
Christian Szegedy, Wojciech Zaremba, Ilya Sutskever, Joan Bruna, Dumitru Erhan,
  Ian Goodfellow, and Rob Fergus.
\newblock Intriguing properties of neural networks.
\newblock {\em arXiv preprint arXiv:1312.6199}, 2013.

\bibitem{goodfellow2014explaining}
Ian~J Goodfellow, Jonathon Shlens, and Christian Szegedy.
\newblock Explaining and harnessing adversarial examples.
\newblock {\em Proceedings of the International Conference on Learning
  Representations (ICLR)}, 2015.

\bibitem{papernot2017arxiva}
Nicolas Papernot, Patrick McDaniel, and Ian~J Goodfellow.
\newblock Transferability in machine learning: from phenomena to black-box
  attacks using adversarial samples.
\newblock {\em arXiv preprint arXiv:1605.07277}, 2016.

\bibitem{Ulicny2016robustness}
Matej Uli{\v{c}}n{\`y}, Jens Lundstr{\"o}m, and Stefan Byttner.
\newblock Robustness of deep convolutional neural networks for image
  recognition.
\newblock In {\em International Symposium on Intelligent Computing Systems},
  pages 16--30. Springer, 2016.

\bibitem{papernot2017practical}
Nicolas Papernot, Patrick McDaniel, Ian~J Goodfellow, Somesh Jha, Z~Berkay
  Celik, and Ananthram Swami.
\newblock Practical black-box attacks against machine learning.
\newblock In {\em Proceedings of the 2017 ACM on Asia Conference on Computer
  and Communications Security}, pages 506--519. ACM, 2017.

\bibitem{Kurakin2017ICLR}
Alexey Kurakin, Ian~J Goodfellow, and Samy Bengio.
\newblock Adversarial machine learning at scale.
\newblock {\em Proceedings of the International Conference on Learning
  Representations (ICLR)}, 2017.

\bibitem{carlini2017towards}
Nicholas Carlini and David Wagner.
\newblock Towards evaluating the robustness of neural networks.
\newblock In {\em Security and Privacy (SP), 2017 IEEE Symposium on}, pages
  39--57. IEEE, 2017.

\bibitem{cisse2017parseval}
Moustapha Cisse, Piotr Bojanowski, Edouard Grave, Yann Dauphin, and Nicolas
  Usunier.
\newblock Parseval networks: Improving robustness to adversarial examples.
\newblock In {\em International Conference on Machine Learning}, pages
  854--863, 2017.

\bibitem{harel2007graph}
Jonathan Harel, Christof Koch, and Pietro Perona.
\newblock Graph-based visual saliency.
\newblock {\em Advances in Neural Information Processing Systems}, pages
  545--552, 2007.

\bibitem{wang2004image}
Zhou Wang, Alan~C Bovik, Hamid~R Sheikh, and Eero~P Simoncelli.
\newblock Image quality assessment: from error visibility to structural
  similarity.
\newblock {\em IEEE Transactions on Image Processing}, 13(4):600--612, 2004.

\end{thebibliography}
